# Stochasticity and Robustness in Spiking Neural Networks

Wilkie Olin-Ammentorp [a], Karsten Beckmann [a], Catherine D. Schuman [b], James S. Plank [c], Nathaniel C. Cady [a]*

[a] State University of New York Polytechnic Institute

257 Fuller Road

Albany, NY 12203, USA

[b] Oak Ridge National Laboratory

1 Bethel Valley Road

Oak Ridge, TN 37830

[c] University of Tennessee, Knoxville

Knoxville, TN 37996

* Corresponding Author

ncady@sunypoly.edu

## Abstract
Artificial neural networks normally require precise weights to operate, despite their origins in biological systems, which can be highly variable and noisy. When implementing artificial networks which utilize analog 'synaptic' devices to encode weights, however, inherent limits are placed on the accuracy and precision with which these values can be encoded. In this work, we investigate the effects that inaccurate synapses have on spiking neurons and spiking neural networks. Starting with a mathematical analysis of integrate-and-fire (I&F) neurons, including different non-idealities (such as leakage and channel noise), we demonstrate that noise can be used to make the behavior of I&F neurons more robust to synaptic inaccuracy. We then train spiking networks which utilize I&F neurons with and without noise and leakage, and experimentally confirm that the noisy networks are more robust. Lastly, we show that a noisy network can tolerate the inaccuracy expected when hafnium-oxide based resistive random-access memory is used to encode synaptic weights.

## 1 Introduction
Artificial spiking networks (ASNs) aim to emulate the abilities that biological neuronal networks demonstrate, including power efficiency, learning capability, and robustness. However, a better understanding of how biological neural networks function, despite being constructed from leaky and unreliable components, may be crucial to enable the construction of complex, large-scale neurally-inspired ('neuromorphic') systems.

This arises mainly from the challenge of creating an artificial version of the biological chemical synapse. Synapses are by far the most prevalent structure within the brain, and each neuron has on average

7,000 synaptic connections. Any neuromorphic system that attempts to emulate this multiplicity requires artificial synapses that are extremely efficient, both in terms of space and energy. While these aspects can be prioritized, it often requires a trade-off with other properties, including precision (the granularity of possible values) and accuracy (how close a value is to what is desired). As a result, many artificial *synaptic devices* that can efficiently encode a weight can also store values, but with some degree of variability. Therefore, an improved understanding of how noise is tolerated (and potentially utilized) within biological neurons and neural networks could enable artificial networks to operate with inaccurate synaptic devices.

As an introduction, we first review the origins and possible roles of randomness within biological neuronal networks. We then overview the field of synaptic devices, with a focus on resistive memories (ReRAM) and the source of their variability. Focusing on the integrate-and-fire neuron model, we examine how its behavior varies as synaptic weights change, and inspect the influence that non-idealities (such as leakage and noise) have on this relationship. From this analysis, we predict that noise can make a neuron more robust against synaptic variability. As an outcome of this analysis, we performed a simulation and implemented an optimization framework to create artificial spiking networks that perform a pole-balancing task. The performance of these networks at their task was re-assessed while using inaccurate weights, and these experimental results confirmed that networks utilizing noisy neurons are more robust.

## 1.1 Noise in Neuronal Networks

Historically, neurons have been investigated and modeled as deterministic systems. Popular models of their action such as the Hodgkin-Huxley and integrate-and-fire models reflect this; as complex as these models can be, they do not model any uncertainty within the neuron's computation (Burkitt, 2006; Hodgkin & Huxley, 1952). This leaves a significant biological aspect of the neuron unaddressed, as it has been long known that both neurons and synapses are subject to noise, and can produce variable results even when given identical stimuli (Branco & Staras, 2009; Calvin & Stevens, 1968).

Variability emerges in neuronal networks through many mechanisms. In chemical synapses, the amount of neurotransmitter emitted to transmit messages to other neurons can change, and furthermore is transported via a random process (diffusion) across the synaptic cleft (Kandel, 2013). This means that despite the well-established notion of a static (and often quite precise) synaptic 'weight', this notion does not truly reflect the reality of a biological synapse.

Neurons themselves are subject to random processes which can influence the nature of their outputs. The random activation of ion-channels can give rise to *channel noise* (local fluctuations in the potential across a neuron's membrane), and this can significantly affect the timing of a neuron's generated action potential. Other factors such as cross-talk between neurons via neurotransmitter release also play a role in influencing the behavior of neurons (Faisal, Selen, & Wolpert, 2009).

The significance, role, and utility of noise in neuronal systems is an area of continuing investigation. Nonetheless, while noise is often viewed as a nuisance which must be minimized, it may play a crucial role in neuronal networks, allowing them to carry out operations such as producing samples from a distribution using a Markov-Chain Monte Carlo (MCMC) method (Buesing, Bill, Nessler, & Maass, 2011). With this biological background in mind, we began to investigate the possible utility of noise within artificial spiking networks.

## 1.2 Synaptic Devices

Neuromorphic systems provide dedicated hardware for executing artificial spiking networks (ASNs), and several large-scale efforts to develop these systems for both commercial and research purposes are underway (Schuman et al., 2017). However, one of the most challenging aspects of creating a neuromorphic system lies in emulating the ability of the synapse. Biological neurons often have thousands of chemical synapses (Yuste, 2015), and to mimic even a small portion of this capability, an artificial synapse must have a footprint which is small as possible, both in terms of energy and space.

Many neuromorphic architectures use digital logic to represent synapses, representing their weights with integral or floating-point values, and storing these values in traditional capacitive or logic-based memories (DRAM/SRAM). But even with nanometer-scale transistors, synapses with several bits of precision can require a large footprint on a neuromorphic circuit, both in terms of area and power, limiting the total number of synapses.

Custom-developed synaptic devices attempt to overcome this difficulty by replacing large digital structures with compact, single-element representations. Potential synaptic devices overlap almost entirely with emerging memory devices, as these two areas require similar functionality. Here, we define an 'ideal' synaptic device as a two-terminal, nanoscale device, which can stably store a value that can be reliably read or modified. The range of distinguishable values this element can store (its precision) and the reliability with which these states can be reached (accuracy) should be as high as possible. Current top candidates for synaptic devices include phase-change memory, ferroelectric memories, and a broad sub-category of devices termed resistive memories (Kuzum, Jeyasingh, Lee, & Wong, 2012; Kuzum, Yu, & Philip Wong, 2013). In this work, we focus on valency-change mechanism (VCM) ReRAM as the potential synaptic device.

## 1.3 Resistive Random-Access Memory (ReRAM)

Resistive random-access memory (ReRAM) is a category which includes a broad range of devices, some of which operate via different physical mechanisms. A strict definition of what constitutes a ReRAM has not been universally agreed upon, but in general, devices which have a two-terminal structure and can utilize a nanoscale reduction-oxidation process to modulate their resistance are classified as ReRAM (Ielmini & Waser, 2015).

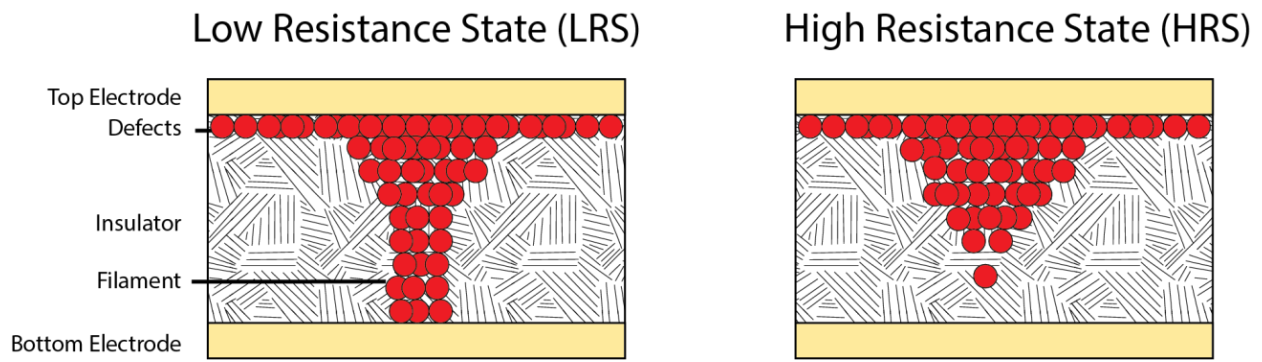

*Figure 1: Illustration of a resistive random-access memory (ReRAM). An insulating oxide (such as hafnium oxide, tantalum oxide, and titanium oxide) is placed between two electrodes. Defects introduced within the film that are mobile under certain thermal and electronic conditions can provide pathways for current to flow between the electrodes. Specialized current pulses can be used to induce conditions that mobilize and re-arrange these defects, programming the device into higher and lower resistance states.*

Most ReRAM's utilize mobile defects within an oxide film to create pathways which can provide a lower-resistance path through an insulator. This conductive pathway can be either created by a conductive bridge constructed from metal ions originating in an electrode, or from oxygen vacancies within the insulator itself (Ielmini & Waser, 2015). Application of an electric field across the insulator can allow these defects to become mobile within the film. The resulting conductive "filament" can be reversibly formed and broken, and also fine-tuned to modulate resistance (Figure 1).

The stochastic transport of vacancies through the film during set/reset means that resistance states are often quite variable. Every programming step changes the distribution of vacancies within the film, creating a unique electronic structure with every operation. This means that accurately programming a device to a specific resistance level can be quite challenging.

Despite its challenges, ReRAM offers many strengths: it can be fabricated from materials which are readily available in integrated circuit fabrication, are highly scalable, require little power, exhibit extremely high endurance, and have clear paths to 3D integration. We have integrated hafnium-oxide based ReRAM within a 65-nm CMOS process flow, and are investigating their performance as a synaptic device in a neuromorphic architecture, the memristive dynamic adaptive neural network array (mrDANNA) (Olin-Ammentorp et al., 2018). Electrical testing data from these ReRAM devices provides a base level of variability which our circuits must be able to withstand.

## 2. The Integrate-and-Fire (I&F) Neuron

One of the most common mathematical simplifications of a biological neuron is the aforementioned integrate-and-fire (I&F) model, and it is used as the basis of artificial neurons within many systems (Burkitt, 2006; Gerstner & Kistler, 2002; Schuman et al., 2017). Essentially, this model describes a neuron as a reservoir which accumulates charge through its synaptic inputs until its voltage reaches a *threshold*. Then, the neuron fires, sending out a signal ('spike') and returning its charge to zero.

A neuron's accumulation of voltage between firing events is described by a differential equation (Equation 1), where $V$ is the potential of the neuron, $C$ is its capacitance, and $I(t)$ is its synapse-injected current at a given time. The neuron is made subject to leakage by adding a term $g_L$, which reduces the voltage by an amount proportionate to its present level. The neuron is made subject to noise by adding a term $\sigma\xi$, where $\sigma$ is the amount of variation and $\xi$ is a Gaussian white noise:

$$C \frac{dV}{dt} = -g_L \cdot V(t) + I(t) + \sigma\xi(t)$$

(Equation 1)

The current $I(t)$ injected into a neuron by its synapses is described by a summation across its inputs (Equation 2). For a neuron with $s$ synapses, the total current injected at a time $t$ is the real-valued weight of each synapse ($w_i$) multiplied by the presence of a spike in its finite spike train $S_i(t)$ at each moment $t$:

$$I(t) = \sum_{i=1}^{s} w_i \cdot S_i(t)$$

(Equation 2)

Each spike train is simply the sum of $n_i$ spikes (with $n_i$ being a natural number), each represented by a Dirac delta function ($\delta$) at unique time offsets $t_{i,j}$ (where $i$ identifies the synapse and $j$ identifies each out of $n_i$ spikes (Equation 3):

$$S_i(t) = \sum_{j=1}^{n_i} \delta(t - t_{i,j})$$

(Equation 3)

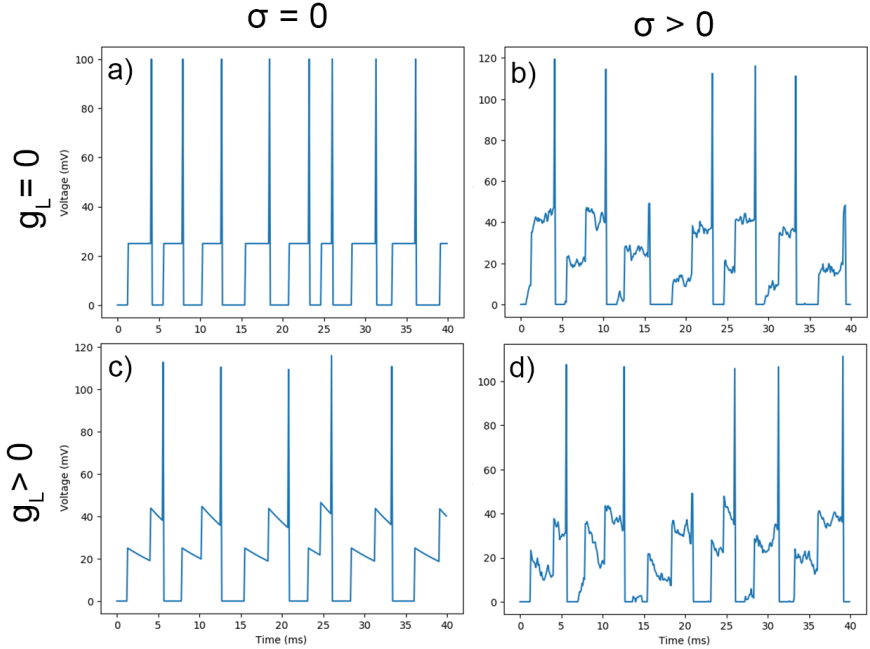

*Figure 2: Plots showing the effects of introducing leakage ($g_L$) and stochasticity (σ) on the spiking output of a single integrate-and-fire (I&F) neuron with current being injected through a single synapse. a) No noise or leakage leads to a simple and rigid integration. b) Noise creates an output which has an unpredictable element. c) Leakage creates sensitivity to the timing of incoming current. d) Leakage and noise together create the most complex behavior.*

Any finite spike train (at least one $n_i > 0$ and every $t_{i,j} < \infty$, Equation 3) applied to a deterministic neuron's (σ = 0, Equation 1) input can be placed into one of two categories: those that caused the neuron to fire, and those that did not. Consequently, for every such non-trivial neuron (defined as having at least one weight $w_i > 0$, Equation 2), we can define a *behavior* as the partition a neuron produces between spike trains into one of two sets: those that excite it to fire, and those that do not. If two neurons with different weights still produce the identical partition of spike trains, we can consider them functionally identical, and state that they have the same behavior. For a network of neurons collectively carrying out a calculation, its results will change if any of its neurons display a different behavior. Thus, if any boundaries between behaviors of an I&F neuron exist, it is important to distinguish where they may lie.

### 2.1 The 'Perfect' Neuron
In digital and simulated spiking neural networks, an idealized neuron with no leakage or noise can be simulated. This 'perfect' neuron carries out a simple operation, which allows us to analytically classify which neurons produce identical behaviors. We then use this understanding to frame the operation of more complex neurons with non-idealities.

By setting $g_L$ and σ (Equation 1) to zero to create a perfect neuron, we obtain the nearly-trivial case where the potential at any time before the first spike is the current number of spikes received by each synapse multiplied by that synapse's synaptic weight (Equation 4), with $\Pi$ representing the Heaviside step function).

$$V_{perfect}(t) = \sum_{i=1}^{s} w_i \sum_{j=1}^{n_i} \Pi(t - t_{i,j})$$

(Equation 4)

In other words, a perfect neuron waits until its synapses have injected it with enough charge to meet or surpass its voltage threshold, at which point it fires. If there is extra charge over the threshold, it is irrelevant and discarded. A consequence of this is that the number of spikes in a in a train needed to make a single-input neuron fire ($f$) is simply the division of its threshold ($\tau$) by the synaptic weight ($w_1$) rounded-up ($ceil$).

$$f = ceil(\tau / w_1)$$

(Equation 5)

If the weight of this single-input neuron is changed, its behavior will only change if the neuron's weight passes a critical value ($w_{crit}$) that integrally divides its threshold. These critical weights can be generated by dividing the threshold by the positive integers ($N^+$), and any two neurons which have a weight on the interval between two critical values will produce identical behaviors. For a single-input perfect neuron, these are the elements of a harmonic series.

$$w_{crit} = \tau/f, f \in N^+$$

(Equation 6)

The condition represented by Equation 6 which separates behaviors for a single-synapse neuron can be extended for an arbitrary number of synapses. Any linear combination of spikes which sum to the threshold defines one firing condition. A matrix can be constructed by defining multiple firing conditions (for instance, two spikes on the first input, and one on the second). When there are as many independent firing conditions are there are synapses, this creates a square matrix ($\boldsymbol{F}$) which represents a constrained system of equations. If it is an invertible matrix, its inverse can be multiplied by a vector of the neuron's threshold ($\boldsymbol{\tau}$) to calculate the weights ($\boldsymbol{w}_{crit}$) which fulfill the firing conditions. Together, these weights and conditions determine the perfect neuron's behavior (Equation 7).

$$\boldsymbol{F} \cdot \boldsymbol{w}_{crit} = \boldsymbol{\tau}$$

$$\boldsymbol{w}_{crit} = \boldsymbol{F}^{-1} \cdot \boldsymbol{\tau}, \quad \boldsymbol{F} \in N^0 \ \& \ |\boldsymbol{F}| \neq 0$$

(Equation 7)

## 2.2 Weight Space

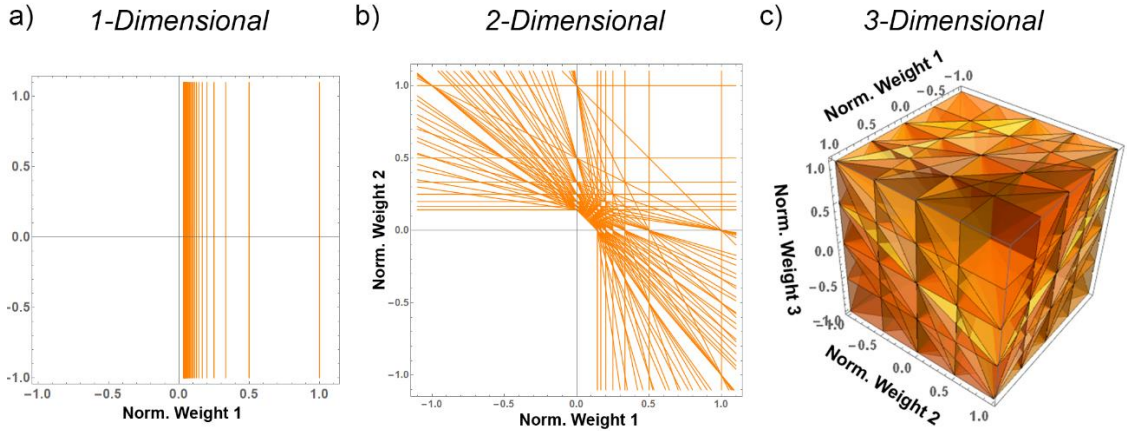

*Figure 3: Hyperplanes in a neuron's weight space describe firing conditions for the neuron, with weights normalized by the neuron's threshold. These form polytopes, each of which maps to distinct output behaviors of a perfect neuron. Approximate representations of these spaces are shown in the first three dimensions (a,b,c). All weights in these representations are normalized to the neuron's threshold. The neuron's output given the same input pattern will only change if a point in this space (representing the weight of incoming synapses) crosses a hyperplane.*

A geometric representation of the boundaries which separate a perfect neuron's behaviors can be deduced from Equation 7. This representation is contained within a hypercube, where each point corresponds to the neuron's threshold-normalized synaptic weight(s). A series of hyperplanes can be drawn through this space, with each plane being defined by a firing condition which defines how inputs can integrally combine to make the neuron fire. These planes partition the space into a set of smaller spaces, or *convex polytopes* (Figure 3). Each polytope corresponds to a unique behavior of a perfect neuron. The intersection of *n* hyperplanes in an *n*-dimensional space correspond to the vector $w_{crit}$ found by solving Equation 7 with the configuration matrix $A$ representing the planes defining the neuron's firing conditions (Figure 4a,b). For a single-input neuron, these hyperplanes correspond to series of points at harmonic values ($\frac{1}{n}$, with $n$ belonging to the positive natural numbers). These boundaries become more complex as the number of weights increases the dimensionality of the weight-space.

Each perfect neuron is represented in weight-space by a point corresponding to the values of its synaptic weights. Perturbations to weights correspond to a movement of this point in weight-space. If this neuron moves far enough to cross over one of the hyperplanes, it enters another polytope, and produces a different behavior. The stability of each behavior is thus proportional to the minimum diameter of its corresponding polytope (Figure 4c,d); if this distance is very small, weight changes will very easily disrupt the neuron's behavior.

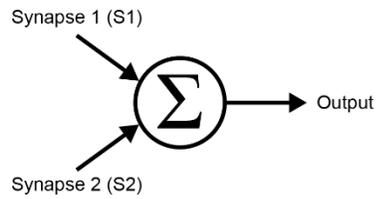
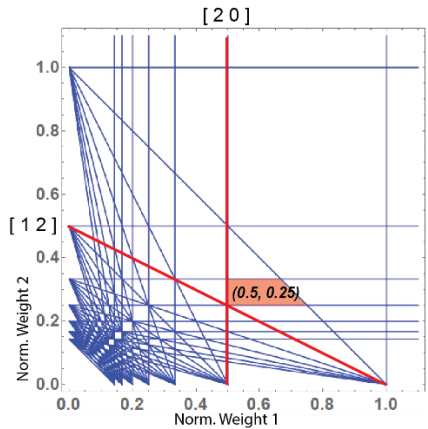
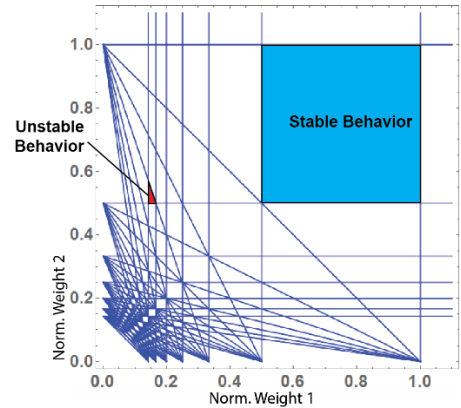

*Figure 4: Illustration of how firing conditions define the weights and behavioral stability of a perfect I&F neuron. An I&F neuron with two synaptic inputs (a) requires two firing conditions to fully specify a system (b), allowing weights that fill that condition to be calculated. The geometric equivalent of this calculation is shown (c), and any weights (here normalized by the neuron's threshold) which fall inside the red quadrilateral will still meet the defined firing conditions; the excess charge accumulated within this area will not be sufficient to cause early firing. Each area bounded by lines corresponds to a unique I&F behavior, and small areas are effectively less stable as a narrower range of weights will produce them (d).*

From this, we can conclude that perturbations to the weights incident on a perfect neuron will not always lead to a change in its behavior. However, if a weight is perturbed across a boundary, the set of inputs which will lead to the neuron firing changes. Consequently, networks that carry out a computation using perfect neurons can be ill-suited to having their weights perturbed, as may occur when synaptic devices with expected programming inaccuracies are used. The network's results may depend on each neuron having one behavior, which may change if the neuron's weights are perturbed outside of that behavior's small region in weight-space. This leads to one potential 'curse of dimensionality' for perfect I&F neurons with multiple synapses; the more synapses that exist, the likelier it is that one of their analog hardware representations will be inaccurate, and induce a different resulting behavior in the neuron.

## 2.3 The Noisy Neuron

In the past, noise in neurons was often considered an environmental nuisance, but there is growing evidence that the stochastic behaviors inherent to biological neurons enables them to collectively perform statistical calculations (Fiser, Berkes, Orbán, & Lengyel, 2010). Here, we consider specifically the impact that noise has on the relationship between a neuron's synaptic weights and its behavior. The voltage of a noisy neuron at a given time can be calculated as the voltage produced by an equivalent perfect neuron (Equation 4), summed with a Wiener process ($W(t)$) representing the noise on the neuron compounded up to a certain point in time (Equation 8).

$$V_{noisy}(t) = \sigma \cdot W(t) + \sum_{i=1}^{s} w_i \sum_{j=1}^{n_i} \Pi(t - t_{i,j})$$

(Equation 8)

Noise which operates directly on the neuron's potential (represented by $\sigma \xi$ in Equation 8), can gradually increase or decrease the potential's value, influencing it so that it may fire (or remain quiescent) when this would not have been the case under noise-free conditions. As a result, the behavior of the neuron is no longer a deterministic function of its weights and input spike trains; the previously strict relationship between these factors is broken. Any spike train has a chance, however remotely, to make the neuron fire, and thus the neuron produces no definite partition between its firing and non-firing spike trains; instead, this classification becomes probabilistic.

For a given spike train, a noisy neuron may fire with fewer or more spikes than would be needed if it were a perfect neuron. When this occurs, we can loosely interpret the noisy neuron as temporarily 'acting' as though it were a perfect neuron, one which has a behavior that could produce the elicited response every time for the applied spike train. Thus, a representation of noisy neurons exists in weight-space as a probability cloud over all possible perfect neurons, with this cloud centered on its neuron's original weights. The amount of noise changes the density of this cloud, as it samples from a range of perfect neuronal behaviors (Figure 5). As noise is reduced to zero, this distribution collapses back to a single Dirac delta at the neuron's position in weight-space. As noise increases, the cloud expands to potentially sample from all possible behaviors. Noise therefore creates a trade-off between how well the neuron can be controlled, and how sensitive its output is to changes in weight.

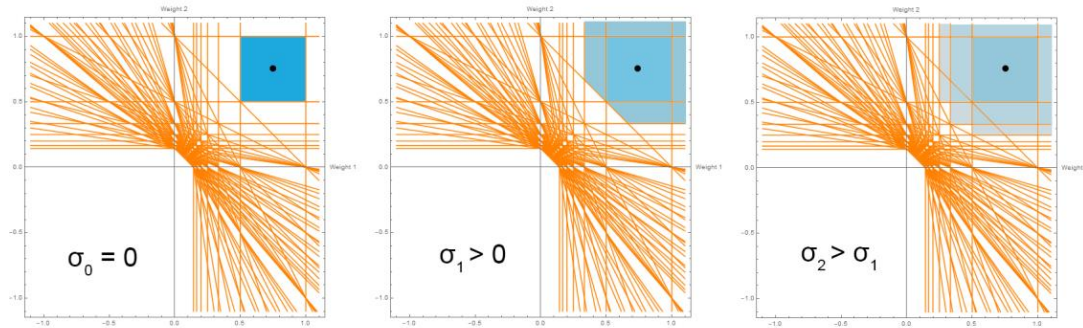

*Figure 5: Illustration of how a noisy neuron expands the range of its potential behaviors. A noise-free neuron only has a single behavior (a). However, the region of behaviors a neuron is likely to sample from expands as noise (σ) increases (b,c), effectively decreasing the impact of the original synaptic weights.*

This probabilistic operation creates a distinct advantage when synaptic weights are imprecise or variable. A perturbation to this cloud in weight-space may cause its center to shift over a boundary between behaviors, but the capability of the neuron to still sample from either side of the boundary means that it can still carry out some of the same processing which it did previously. Additionally, networks constructed from noisy neurons must already have the capability to withstand variable, non-deterministic computation for them to operate. In contrast, networks without noise may learn to utilize a narrow set of weights and behaviors which may catastrophically fail when perturbed.

### 2.4 The Leaky Neuron

Leaky neurons introduce the complication that the charge in the neuron can escape, making its voltage decay over time until its returns to equilibrium. This 'imperfection' adds a crucial temporal element to the neuron, displayed in the analytic solution to the leaky I&F neuron (Equation 9). As each spike arrives at a synapse, it initially contributes its full weight ($w_i$), but this charge exponentially decays away as the system's time ($t$) continues past the spike's arrival time ($t_{i,j}$). This allows the neuron to effectively filter signals based on their time of arrival, and due to this time-dependency, the exact partition of spike trains into firing and non-firing categories (as we established for the perfect I&F neuron) becomes much more complex.

$$V_{leaky}(t) = \sum_{i=1}^{s} w_i \sum_{j=1}^{n_i} \exp[-g_L(t - t_{i,j})] \cdot \Pi(t - t_{i,j})$$

(Equation 9)

For any change in weights to a leaky neuron which operates in continuous time, we can find a spike train which changes its excitatory/non-excitatory classification after the weight change. Due to the decay term in Equation 9, we can construct a spike train with a specific temporality that brings the neuron to the brink of firing, and even a tiny change to weights changes the neuron's response to this spike train across the firing threshold. As a result, leaky neurons have no distinct classes of behavior which can be mapped to the geometry of the firing conditions within weight-space; each leaky neuron with unique weights has a unique behavior. It is worth considering, however, that in clocked spiking networks, the discretization of time in which it operates decreases the cardinality of the set of possible spike trains. Effectively, it is possible for two distinct leaky neurons in a clocked system may be equivalent, given the more limited set of spike trains on which they actually operate.

Directly inferring the effects of random changes to a leaky neuron's weights is difficult, but we may conclude from each leaky neuron's uniqueness that there is no reason to believe that introducing leakage into a perfect neuron would provide a benefit in terms of resilience to weight perturbation, and it may actually increase its sensitivity.

## 2.5 The Noisy, Leaky Neuron

Combining the two non-idealities considered here (leakage and noise) creates the neuron with the most complex behavior. This neuron's output is both non-deterministically related to its weights, and dependent on the temporality of its inputs. Based on the strength of the noise and the time scale of the leakage, it is possible that this type of neuron could display either more or less robustness, based on the interplay of these two effects. This compounded effect, however, makes it challenging to analyze analytically. We can determine that due to the leakage, each unique set of weights can produce a unique behavior, and the noise will make this behavior probabilistic. The robustness of this neuron to weight perturbation likely depends on how these effects interact, and we believe this is more easily established empirically than analytically.

Table 1: A summary of the types of neurons being investigated and their properties.

| Neuron Type | $g_L$ | $\sigma$ | Deterministic | Temporal Characteristics | Predicted Robustness |
|---|---|---|---|---|---|
| Perfect | 0 | 0 | Yes | Accumulates | Low |
| Leaky | > 0 | 0 | Yes | Time-dependent accumulation | Low |
| Noisy | 0 | > 0 | No | Variable accumulation | High |
| Leaky / Noisy | > 0 | > 0 | No | Time-dependent, variable accumulation | Moderate |

## 3. Experimental Measurement of Robustness

Two experiments were carried out to investigate whether our conclusions from this theoretical analysis provide a valid basis for predicting the ability of neurons and neural networks to be robust to synaptic inaccuracy. First, simulations of single spiking neurons under perfect and noisy conditions were carried out to verify our predictions of where their behavioral boundaries lie. Second, sets of spiking neural networks were trained to carry out a pole-balancing task. Each set utilized neurons with different non-idealities, and the weights within this network were progressively perturbed away from their original values. The effect that this perturbation had on network performance was measured to verify our prediction of overall network robustness with different non-idealities.

### 3.1 The Effect of Noise on Individual Neurons

Single integrate-and-fire neurons defined by the differential equation shown above (Equation 1) were simulated using the Brian2 package (Stimberg, Goodman, Benichoux, & Brette, 2014). When the noise variability ($\sigma$) was greater than 0, the stochastic differential equation was solved using the Milstein method (Milstein, Platen, & Schurz, 1998). These neurons had two synaptic inputs, and random spike trains were applied to these inputs and the neuron's firing was observed. The firing output of the neuron using all synaptic weights within a range below and equal to the neuron's threshold (50 mV) was recorded. For each set of synaptic weights, it was recorded whether the output was different than that of its nearest neighbors with higher or lower weight values. This provided an experimental image of where boundaries exist in the neuron's behavior.

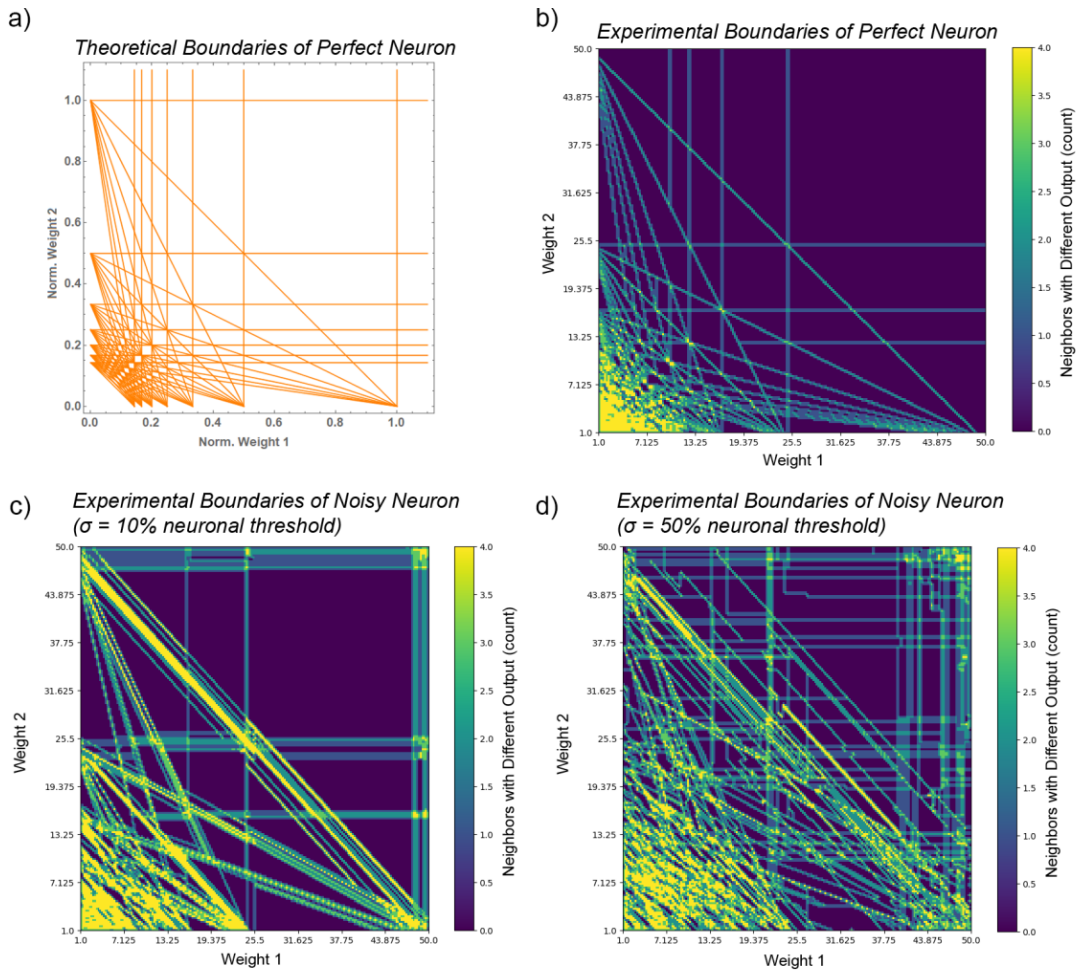

*Figure 6: Experimental results to detect the boundaries between behaviors in different types of I&F neurons using excitatory weights. When there is no noise, the theoretical boundaries (a, with weights normalized by the threshold) almost exactly match the experimental boundaries (b, threshold of 50 mV), and likely only diverge due to the finite nature of the testing used. However, when noise is introduced, the neuron can stochastically 'act' with a different behavior. The boundaries between behaviors become convoluted across different regions depending on the level of noise used (σ = 10% of threshold value (c), σ = 50% threshold value, (d)).*

The experimentally-mapped boundaries of a perfect I&F neuron almost exactly matched those predicted by the theoretical analysis (Figure 6a,b). Discrepancies between the figures where predicted boundaries were not seen are likely due to the finite testing applied to the simulated neuron, which did not create all possible neuronal behaviors.

Adding noise to the neuron's potential could effectively allow it to act with different behaviors that require higher or lower weights, and this was reflected in the experimental mapping of a noisy neuron's boundaries, which appear slightly displaced from their original positions when lower amounts of noise are included (Figure 6c). This resulted in a boundary map that resembled the original image, but convoluted over new points. As noise continued to increase, boundaries were translated further and further away, and the original structure of the boundaries became unrecognizable (Figure 6d). These results support our hypothesis that with no noise, the behavior of an I&F neuron changes discontinuously as it encounters rigid boundaries. Noisy neurons can change behaviors more continuously, shifting their responses to incoming spike trains gradually and probabilistically as their synaptic weights change.

## 3.2 Fitness and Stability of Neural Networks Evolved with Noise or Under Ideal Conditions

An evolutionary optimizer (EO) was used to evolve three-dimensional spiking networks constructed from integrate-and-fire neurons (Plank et al., 2018; Schuman, Birdwell, & Dean, 2015). These networks were optimized to keep a pole on a movable cart upright for a certain period of time. All inputs to and outputs from the network are solely in terms of spikes, and the network carries out a spike-based computation. Two spiking outputs vote to either move the cart left or right for each simulation period (corresponding to 20 ms of simulated time), and through this operation the network can keep the cart balanced. The fitness of each spiking network is the number of cycles over which it can keep the pole balanced, up to a limit of 15,000 simulated cycles (5 minutes of simulated balancing).

Due to the limitations of our spiking simulator as an event-based system, an approximation was made to allow networks to have stochastic behavior. As a result, it does not compute true compounded noise as a continuous process (as in Equation 1), but instead has a probability to spike every time its potential changes (Equation 10, after an Arrhenius escape rate (Plesser & Gerstner, 2000)). We believe this creates a sufficient approximation of the noisy neuron to estimate whether it can improve robustness.

$$P(fire|v) = Exp\left(-\frac{\tau - v}{\tau \cdot \sigma_a^2}\right)$$

(Equation 10)

Networks were created for each type of neuron being investigated: perfect, leaky, noisy, and leaky/noisy. For leaky networks, a time constant ($T = 1/g_L$) of 50 cycles was used, and for noisy networks, a deviation ($\sigma_a$) of 0.60 was used. This created a decay on a period which is relevant to incoming signals without causing them to disappear before they can be processed, and a level of noise which makes the firing rate of a regularly-driven neuron increase approximately linearly with an increasing synaptic weight.

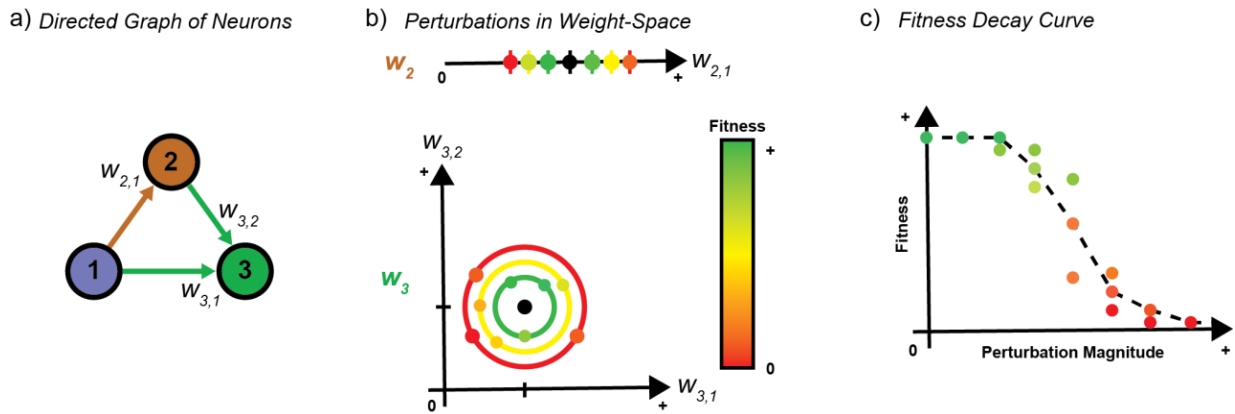

Figure 7: Illustration of how the sensitivity of each network to perturbation was measured. a) The weights of each neuron in the network are represented as single points in weight-space. These points are moved in a random direction to a certain distance, and the fitness at these new points is re-evaluated multiple times (b). The fitness of the networks as their weights stray increasingly from the initial values is plotted, and the median value is taken as the trend (c).

To test the sensitivity of each network with different neuronal non-idealities to perturbation, the weights of synapses incident on each of its neurons were represented as vectors (equivalent to points in each neuron's weight-space, Figure 7a). Vectors were added to each neuron's weights with random direction, but constant magnitude. This operation moved the representation of a neuron in weight-space to a distance a constant radius away (Figure 7b). The fitness of the network with these perturbed weights was re-calculated, with 1 trial for non-noisy networks, and the median of 20 trials taken for noisy networks (due to their stochastic operation).

The original weights of the network were then perturbed again in a different random direction. Ten samples were taken at each magnitude of perturbation, with the median value representing the fitness of the network under these changes (Figure 7c). By increasing the perturbation magnitude and repeating this process, we create a curve which illustrates how quickly a network's performance degrades as its neurons' synaptic weights are perturbed away from their original values.

We evolved 34 perfect, 32 leaky, 14 noisy, and 12 leaky/noisy networks to perform the pole-balancing task. All non-noisy networks demonstrated balancing to a full 15,000 cycles. However, this was more difficult to achieve for the noisy networks, as our optimizer is currently configured with the assumption that fitness is deterministic, and as a result its solution networks would often display inconsistent performance. Therefore, noisy networks were only accepted if their median balancing time over 20 trials exceeded 12,000 cycles out of the maximum trial time of 15,000 cycles when tested on the same conditions. Noisy networks without leakage were more often able to pass this threshold, and showed very high mean performance (Figure 8a). The combination of two non-idealities in the leaky/noisy networks made them much more difficult to optimize, and none displayed consistently high performance (Figure 8b). Because of the challenges in optimizing the non-deterministic networks, fewer successful networks were produced compared to the number of deterministic networks generated using roughly the same computational resources.

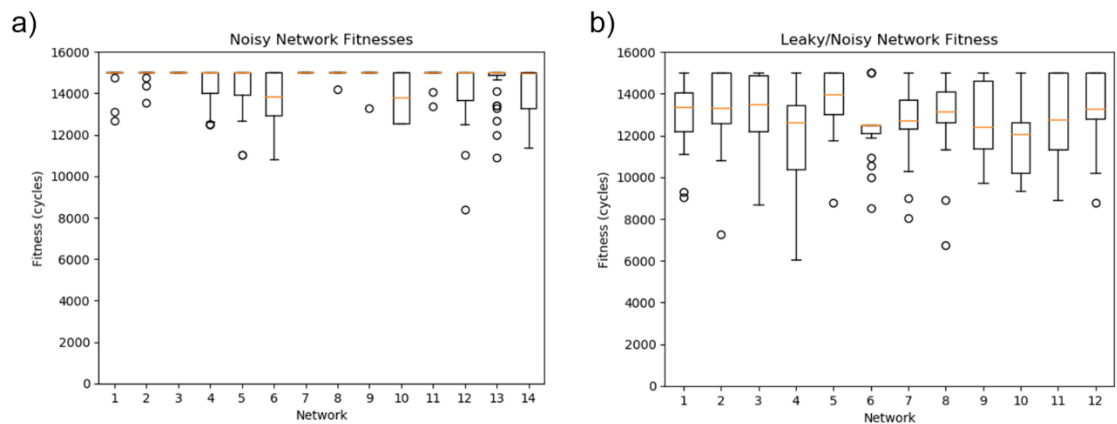

*Figure 8: The range of fitness values produced by spiking networks using a noisy neuron. Due to their stochastic behavior, each network produced a range of fitness values. Consistently high fitness was demonstrated more often for networks with only noise (a), as opposed to both noise and leakage (b). Box-and-whisker plots used in this work place the box around the interquartile range (IQR), with yellow line indicating the median. Whiskers show the range of data which exists past the first or third quartiles but within 1.5 times the IQR from these limits. Data exceeding these limits (outliers) are plotted as circles.*

The collections of networks evolved under each condition were perturbed over a range which made fitness decay to low or zero fitness values. A cursory inspection of the curves produced (Figure 9) shows that networks that utilized a noisy neuron could maintain a high fitness at their task under large amounts of perturbation (approximately 20% perturbation in a normalized weight space). This is in contrast to networks using non-noisy neurons, which appear to begin losing their capabilities even under small perturbations to their weights (2% perturbation).

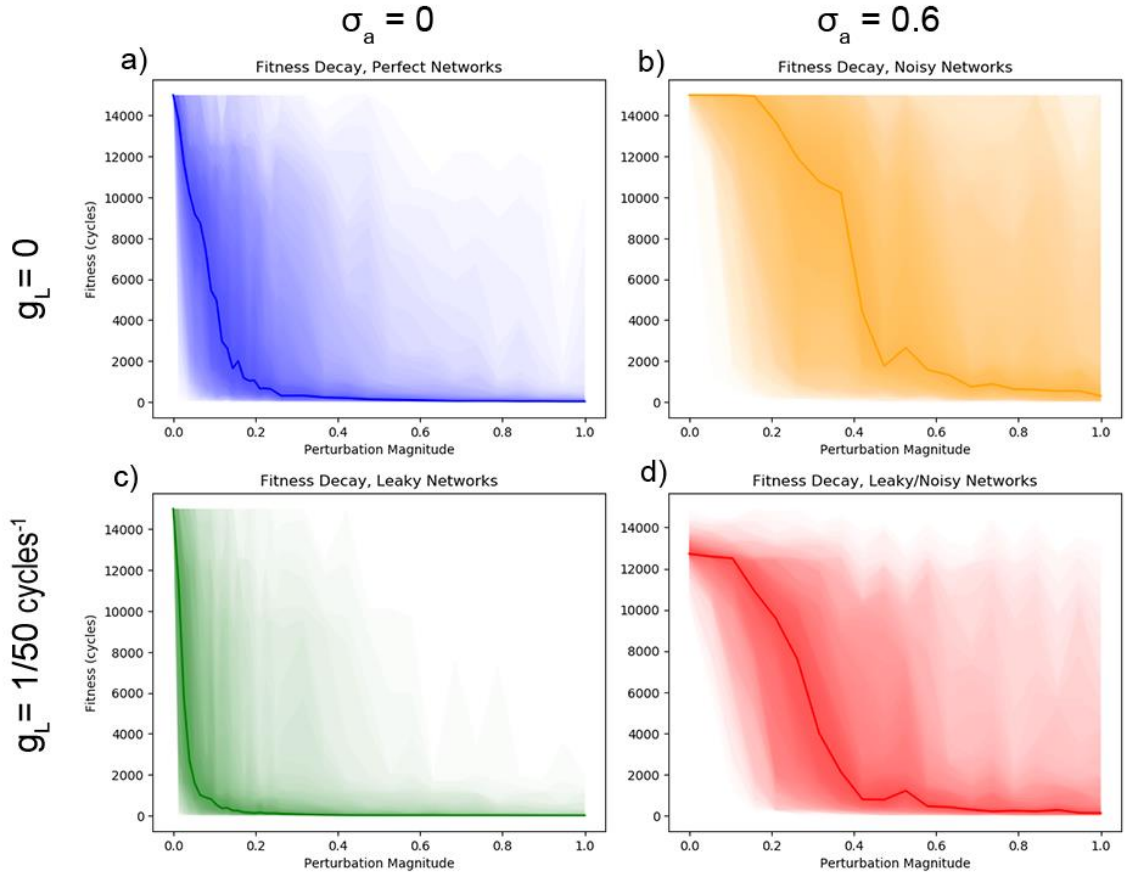

*Figure 9: The effect which perturbing the weights of each neuron in a spiking network has on its ability to carry out a pole-balancing task, for neurons with different types of non-idealities (leakage ($g_L$) and noise ($\sigma$)). The density of color in each plot corresponds to the number of networks in that region, and the solid line plots the median fitness value at each perturbation level. Perturbation magnitude is equivalent to a distance within a normalized weight space. Spiking networks using noisy neurons (b,d) appear to decay less quickly with perturbation than networks using non-noisy neurons (a,c).*

To quantify and test this hypothesis that the noisy neurons are more resilient to perturbation, we calculate the magnitude of perturbation required to reduce each network's median performance to 50% of the maximum possible fitness value. By comparing the distribution of this metric for networks using each type of non-ideality, we can conclude with high confidence that this distance is greater for networks using noisy neurons (Figure 10). Additionally, introducing leakage does not appear to introduce a significant change in the magnitude of perturbation required to reduce network fitness to 50%.

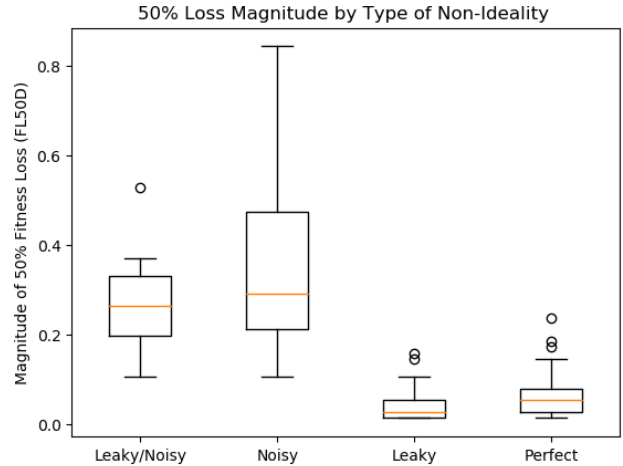

*Figure 10: The distribution of magnitudes required to reduce networks to 50% of the maximum possible fitness level is plotted. This magnitude is significantly greater when using noisy neurons (unequal variances t-test, p < 0.1%), demonstrating that noisy networks are less sensitive to weight perturbation.*

### 3.3 ReRAM-based Synaptic Devices for Neural Networks

Having established theoretical and experimental evidence that noisy neurons can be more robust to inaccurately-programmed weights, we wished to investigate whether this robustness is sufficient to withstand the variabilities and limitations expected in real synaptic devices. One of the limitations of synaptic devices is that within their dynamic range, variability in programming will limit the number of distinguishable states that can be reliably achieved. Each of these states may also still contain a variability.

To enable synaptic devices as integrated circuit components, we have previously established hybrid CMOS-ReRAM one transistor - one ReRAM (1T1R) cells using a 65nm CMOS baseline process (Olin-Ammentorp et al., 2017). Using a pulse-based system to gradually reset devices from the low-resistive state (LRS) up to a desired value (with read-verification), we can program intermediate states between a 2.75 kΩ low resistance state (LRS) and 11.5 kΩ high resistance state (HRS). We have also implemented ReRAM as synaptic devices in simple neuron circuits, in which a pair of 1T1R cells is used to represent a single synapse. This is due to the asymmetric programming behavior of our hafnium oxide based ReRAM (Beckmann, Holt, Manem, Van Nostrand, & Cady, 2016). To enable both potentiation and depression of the synapse, two 1T1R cells are used, so that one can be used to increase (and the other decrease) the synaptic weight. Generally, one device is referred to as being 'excitatory' and the other 'inhibitory' (Figure 11a) (Schuman et al., 2019). When using 2 devices in this differential-pair configuration, *n* analog device levels yield *(2n-1)* weight levels, giving 7 possible synaptic weight levels from a 4-level ReRAM element. Some of these values can be represented in multiple ways using the pair of 1T1R cells (Figure 11b), and each of these representations is sampled equally when establishing the distributions for variability at each weight level.

For each representation, a sample of a ReRAM element's electrical resistance at that level is sampled from a normal distribution determined by testing (Table 2). These states were selected to provide a coverage across the ReRAM's achievable dynamic range, and achieve mean programmed values which were evenly spaced as possible, preventing weights from widely differing from the programming value.

Table 2: Distributions of resistance values for the 4 selected resistance states of the HfOx 1T1R cell.

| Resistance Level | Targeted Value (ohm) | Mean, Programmed Value (ohm) | Std. Dev, Programmed Value (ohm) |
|---|---|---|---|
| 1 | 2750 | 2790 | 41.3 |
| 2 | 5000 | 5610 | 425 |
| 3 | 7750 | 8380 | 479 |
| 4 | 10500 | 11300 | 617 |

The distribution of synaptic weights achieved by these ReRAM representations was calculated by drawing 1,000 samples at each of the 7 weight levels (Figure 11c,d). We found that the pulsed reset/verify programming method is crucial to achieving accurate weights, but with this method, distinguishable distributions of weight values can be achieved.

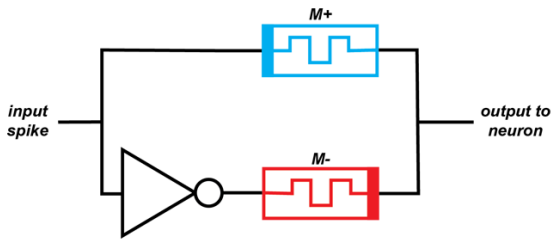
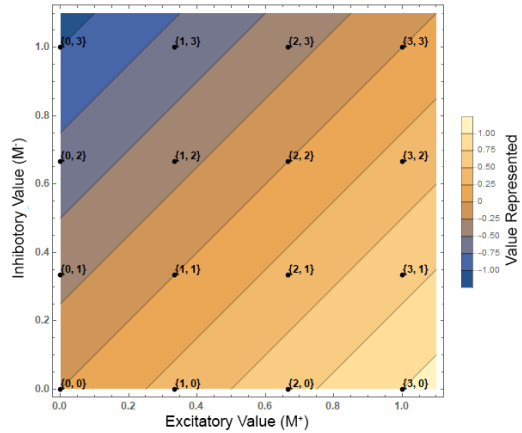
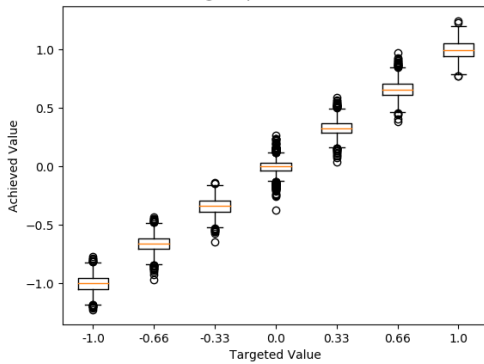
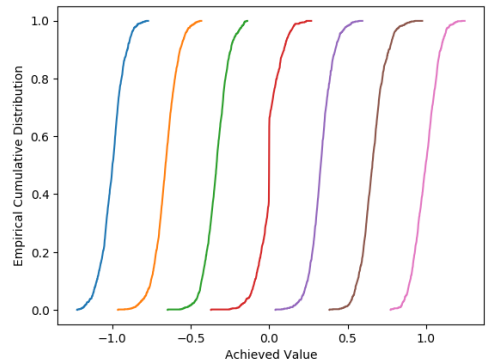

Figure 11: A pair of excitatory/inhibitory ($M^+/M^-$) ReRAM devices (a) is used to represent a synapse so that weights can be evenly modulated up or down even when the synaptic device has asymmetric programming characteristics. In this 'twin synapse,' each value can have multiple representations (b), and the representations for each value given a 4-level device are shown. Assuming an equal probability that each representation is used, the variability of the 7 resulting weight values using pulse-programmed HfOx ReRAM is shown (c,d).

## 3.4 The Effects of ReRAM Stochastic Behavior on Noisy vs. Ideal Neural Networks

Using networks that were evolved in Section 3.2, we sub-selected two exemplary networks which could also withstand a reduction to 7 synaptic weight levels (4 device levels). These networks maintained full performance when weights were rounded to the nearest representable level, allowing them to be fully adapted by synapses that could perfectly represent these 7 levels. One of these networks utilized perfect I&F neurons, and the other used noisy I&F neurons. The variability of the weights in these networks was gradually raised to the level expected using fabricated HfOx 1T1R cells (from our laboratory), with network fitness at each step sampled 100 times. This established the range of performance degradation a perfect or noisy I&F network might exhibit when transferred to hardware using ReRAM as synaptic devices.

As expected, the behavior of the network using perfect I&F neurons became irregular and the median fitness value degraded, while the weight variability increased (Figure 12a). In contrast, the network using noisy I&F neurons maintained a median performance value at the maximum value even under the full variability levels expected from a real ReRAM device (Figure 12b). We believe that this provides evidence that the view of neurons as stochastic computing elements is a powerful perspective which can be used to construct robust networks that can utilize relatively inaccurate or stochastic elements, such as most ReRAM devices that have been reported to date. This provides motivation to consider more stochastic designs for future neuromorphic systems and optimization/simulation frameworks.

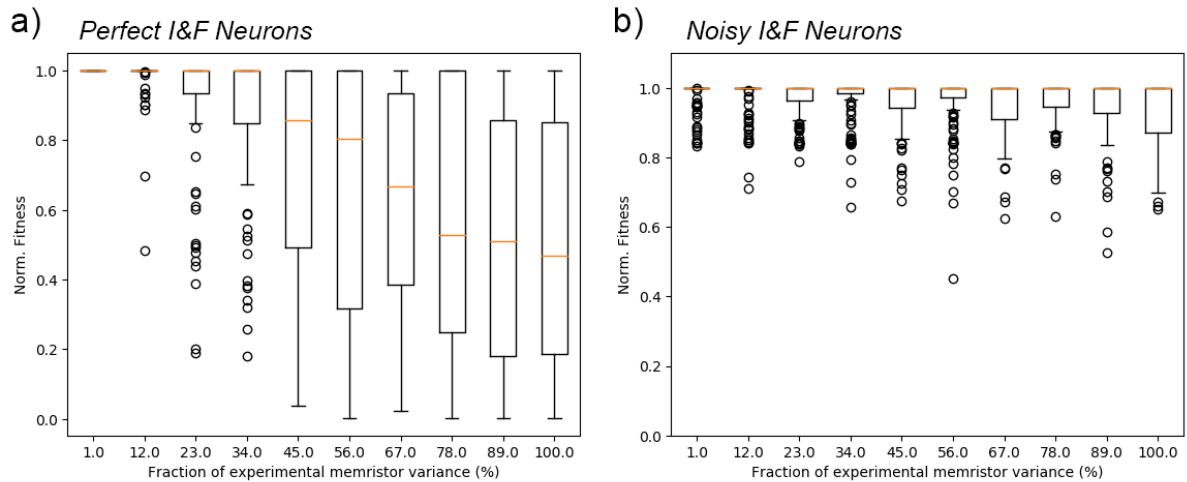

*Figure 12: The performance of networks using perfect (a) and noisy (b) I&F neurons carrying out a pole-balancing task as synaptic variability increases to levels expected under ReRAM-based implementation. Fitness values are normalized to the maximum level (15,000 cycles). As a result of the perfect neuron's fragility, the perfect network's fitness becomes variable when inaccurate memristive weights are utilized. In contrast, performance variability may slightly increase for the noisy network using memristive weights, but its median performance remains at the maximum level.*

## 5. Conclusions

Compact representations of synaptic weights will likely be necessary in order to create large-scale neuromorphic architectures with many inputs and outputs per neuron. However, this will likely come at the cost of accuracy. This may be reasonable, however, given the fact that biological synapses themselves are quite imprecise. To generate future neuromorphic systems capable of incorporating inaccurate synaptic devices, we require a better understanding of how biological systems tolerate this inaccuracy.

From a theoretical analysis of integrate and fire neurons, we have shown that neurons with no non-idealities (such as leakage and internal noise) undergo discontinuous changes in behavior as their weights cross a critical boundary value. However, when noise is introduced to the neuron, its output is no longer strictly dependent on its weights, and its behavior can more slowly diverge from its previous state as its synaptic weights are perturbed. From this, we predict that spiking networks using noisy neurons can be more resilient to synaptic inaccuracy.

This conclusion is validated by two experimental studies. First, simulations of spiking neurons confirmed that boundaries between behaviors exist where expected. Second, collections of spiking neural networks evolved for a pole-balancing tasks were studied. The synaptic weights of these networks were perturbed from their optimal values, and their fitness at the task was re-evaluated. We confirmed that the magnitude of perturbation required to reduce the fitness of noisy networks to 50% of the peak value is significantly greater than for networks using non-noisy neurons.

Lastly, we examine networks operating with a level of synaptic variability equivalent to an amount derived from experimentally-measured resistive random-access memory (ReRAM). We find that when noisy integrate-and-fire neurons are used over noise-free neurons, the resulting networks can much more often tolerate the expected ReRAM variance. We believe this provides motivation to consider stochasticity as an important element in future neuromorphic designs incorporating analog synaptic devices.

**Declaration of Interest:**

Declarations of interest: none

**Funding:**

This work was supported by AFRL grant FA8750-16-1-0063.